\pdfoutput=1

\documentclass[11pt]{article}

\usepackage[]{emnlp2021}

\usepackage{times}
\usepackage{latexsym}

\usepackage{microtype}

\usepackage{hyperref}
\usepackage{url}
\usepackage{latexsym}
\usepackage{soul}
\usepackage{caption}
\usepackage{graphicx}
\usepackage{amsmath}
\usepackage{amsthm}
\usepackage{booktabs}
\usepackage{algorithm}
\usepackage{algorithmic}
\usepackage{multirow}
\usepackage{graphicx}
\usepackage{amssymb}
\usepackage{amsfonts}
\usepackage{todonotes}
\usepackage{tabularx}
\usepackage{arydshln}
\usepackage{mathtools,nccmath}
\usepackage{wrapfig}
\usepackage[utf8]{inputenc}
\usepackage[utf8]{vietnam}
\usepackage[english]{babel}
\usepackage{enumitem}

\title{Automatic Post-Editing for Vietnamese}

\author{
    Thanh Vu$^{1}$ \and Dai Quoc Nguyen$^{2}$\thanks{\ \ Most of the work was done before two authors joined Oracle.}\\
    $^1$Oracle Digital Assistant, Oracle, Australia\\ 
    $^{2}$Oracle Labs, Oracle, Australia\\
    \tt{\normalsize $^{1}$thanh.v.vu@oracle.com; $^{2}$dai.nguyen@oracle.com}
}

\begin{document}

\maketitle

\begin{abstract}
Automatic post-editing (APE) is an important remedy for reducing errors of raw translated texts that are  produced by machine translation (MT) systems or software-aided translation. In this  paper, we present a systematic approach to tackle the APE task for Vietnamese. Specifically, we construct the first large-scale dataset of 5M Vietnamese translated and corrected sentence pairs. We then apply strong neural MT models to handle the APE task, using our constructed dataset. Experimental results from both automatic and human evaluations show the effectiveness of the neural MT models in handling the Vietnamese APE task. 

\end{abstract}

\section{Introduction}
Recent research has placed significant advancements for automatic machine translation \cite{WuSCLNMKCGMKSJL16,vaswani2017attention,barrault2019findings}.
The high-quality MT output has been widely adopted by professional translators into their translation workflow to save time and reduce translation errors \cite{zaretskaya2016comparing}.

Translating Chinese novels to Vietnamese is an important task. In the last ten years, there are about 30K Chinese novels  describing fiction stories, that are available in Vietnamese  with $\sim$ 80K  active readers and $\sim$ 600K novel chapter views daily from the three most popular Vietnamese websites for reading  novels.\footnote{\url{https://truyencv.com}}\footnote{\url{https://truyenyy.com}}\footnote{\url{https://truyen.tangthuvien.vn}}
But, translating the Chinese novels to Vietnamese is still challenging.
The reason is that in fact, readers prefer reading the novels translated using the traditional language style rather than the modern language style used in news articles (e.g. using 
``\textbf{tiểu nữ nhi}''$_{\text{\textbf{little girl}}}$ instead of ``\textbf{cô bé}''$_{\text{\textbf{little girl}}}$).
Note that current general-purpose MT systems (e.g., Google Translate), trained on modern language style-focused bilingual corpora, cannot satisfy the reader preference.

The well-known workflow/guideline used for translating the Chinese novels to Vietnamese consists of three steps:\footnote{\url{http://www.tangthuvien.vn/forum/showthread.php?t=142168&page=2}}

\begin{itemize}[leftmargin=*]
\setlength\itemsep{-1pt}
    
\item In the first step, the Chinese text is converted into Sino-Vietnamese  (i.e. Han-Viet)\footnote{\url{https://en.wikipedia.org/wiki/Sino-Vietnamese_vocabulary}} text using a specialized software, such as TTV Translator.\footnote{\url{https://play.google.com/store/apps/details?id=vn.tangthuvien.ttvtranslate&hl=en_AU}.}
    
\item In the second step, the Sino-Vietnamese text is further smoothed by replacing predefined Sino-Vietnamese phrases by dictionary-based Vietnamese phrases. 
The core content of the Vietnamese text generated as the output of the second step---namely software-aided \textbf{translated} text---can be generally understood by frequent readers who are familiar with reading the translated text. 
Note that the translated text does not fully follow the Vietnamese grammar and vocabulary, thus making it hard for new readers (and even fairly often for the frequent readers) to understand details of the text content. 

\item In the final step, the translated text is manually edited and polished following Vietnamese vocabulary and grammar. Here, we refer to the text generated as the output of the  final step as the human-\textbf{corrected} text  that  can be accessed easily by readers with different reading levels. 

\end{itemize}

Note that the final editing step is very time-consuming due to the large amount of human-manual work.
Thus automatic post-editing (APE) might be involved in this final step, helping to reduce the human effort in editing the translated text \cite{tatsumi2010post}. 
To the best of our knowledge, there is no previous study on APE for Vietnamese.

In this paper, we formulate the APE problem for Vietnamese as a  monolingual translation task. 
We first construct a large-scale dataset consisting of  translated and corrected sentence pairs. 
We then use our dataset to train a state-of-the-art neural MT model to automatically post-edit the translated sentences, and compare these models under various settings. 
Our contributions are summarized as:

\begin{itemize}[leftmargin=*]
\setlength\itemsep{-1pt}
    \item We are the first to tackle the APE task for Vietnamese to automatically improve the quality of the Vietnamese translated text of Chinese novels. We create a large-scale dataset of 5M translated and corrected sentence-level pairs extracted from 99.5K translated and corrected chapter-level pairs from 183 novels.
    
    \item We empirically evaluate neural MT models using our dataset, including a fully convolutional model \cite{gehring2017convolutional},  ``Transformer-base'' and ``Transformer-large'' \cite{vaswani2017attention}. We compare these models under automatic- and human-based evaluation settings as well as in-domain and out-of-domain schemes. 
\end{itemize}

\section{Our dataset}
This section presents our large-scale dataset for the Vietnamese APE task.

\subsubsection*{Dataset construction} 
In almost all cases, the original Chinese novels are not publicly available to the readers of the Vietnamese websites for reading novels, thus \textit{we cannot access those Chinese novels' texts}.
Of 30K Chinese novels available in Vietnamese, there are currently only 283 novels available in {both} Vietnamese translated and corrected texts. We crawl all of those 283 novels. 
There is a ground-truth chapter-level alignment between translated and corrected chapter-level pairs from each of the 283 novels. 
We randomly sample from each novel 5 pairs of translated and corrected chapters and employ three annotators to manually evaluate the sampled chapters' editing quality on a 5-point scale. We select the top 183 novels having the highest average points 
over their sampled chapters to be included in our dataset. 

We use all translated and corrected chapter-level pairs from the top 183 novels, i.e. a total of 99.5K chapter-level pairs. We then use RDRSegmenter \cite{nguyen-etal-2018-fast} from VnCoreNLP \cite{vu-etal-2018-vncorenlp} to segment each chapter text into individual sentences.  
In each chapter, to align the translated and corrected sentences, we compute an alignment score $\alpha = \dfrac{2 \times |I|}{|T| + |C|}$, where $|T|$ and $|C|$ denote the numbers of tokens 
in the translated and corrected sentences, respectively, while $|I|$ denotes the size of the intersection between them. Our sentence alignment process has two phases: 

\begin{itemize}[leftmargin=*]
\setlength\itemsep{-1pt}
\item In the first phase, we  align every translated and corrected sentence pair  with a score $\alpha >= 0.75$, i.e. alignment mode 1--1. 
\item In the second phase, for the remaining sentences, using a threshold $\alpha >= 0.5$, we only consider two alignment modes 1--2 and 2--1  for one translated sentence aligning two adjacent corrected sentences and two adjacent translated sentences aligning one corrected sentence, respectively.\footnote{We concatenate two adjacent sentences into a single one.} 
\end{itemize}

The alignment modes 1--1, 1--2 and 2--1 account for about 98\% of the validation set.\footnote{We do not include the remaining 2\% unaligned sentences into our dataset.} In the end, our dataset consists of 5M (i.e. 5,028,749)  translated and corrected sentence-level pairs in Vietnamese.

\begin{table*}[!t]
\centering
\resizebox{14.75cm}{!}{
\begin{tabular}{l|l|l|l|l|l|l}
\hline
\multirow{2}{*}{Item} & \multicolumn{2}{c|}{Training set}  & \multicolumn{2}{c|}{Validation set} & \multicolumn{2}{c}{Test set}\\
\cline{2-7}
\ & Translated & Corrected & Translated & Corrected & Translated & Corrected\\
\hline
\#chapters(\#novels) & \multicolumn{2}{c|}{92.2K (183)}& \multicolumn{2}{c|}{2.5K (183)}& \multicolumn{2}{c}{4.8K (183)}\\\hline
\#sentences & \multicolumn{2}{c|}{4.65M} & \multicolumn{2}{c|}{126.7K} & \multicolumn{2}{c}{ 248.0K} \\
\hline
\#tokens & 152.1M & 143.7M & 4.1M & 3.9M & 8.1M & 7.6M \\
\hline
\#tokens/sentence & 32.7 & 30.9 & 32.7 & 31.0 & 32.6 & 30.8 \\
\hline
\end{tabular}
}
\caption{In-domain statistics of our dataset.}
\label{tab:in-domain:statistics}
\end{table*}

\begin{table*}[!t]
\centering
\resizebox{14.75cm}{!}{
\begin{tabular}{l|l|l|l|l|l|l}
\hline
\multirow{2}{*}{Item} & \multicolumn{2}{c|}{Training set}  & \multicolumn{2}{c|}{Validation set} & \multicolumn{2}{c}{Test set}\\
\cline{2-7}
\ & Translated & Corrected & Translated & Corrected & Translated & Corrected\\
\hline
\#chapters(\#novels) & \multicolumn{2}{c|}{91.5K (128)} & \multicolumn{2}{c|}{2.8K (28)} & \multicolumn{2}{c}{5.1K (27)} \\\hline
\#sentences & \multicolumn{2}{c|}{4.66M} & \multicolumn{2}{c|}{120.1K} & \multicolumn{2}{c}{245.6K} \\
\hline
\#tokens & 151.3M & 143.0M & 4.1M & 3.8M & 8.9M & 8.4M \\
\hline
\#tokens/sentence & 32.5 & 30.7 & 33.7 & 31.6 & 36.3 & 34.2 \\
\hline
\end{tabular}
}
\caption{Out-of-domain statistics of our  dataset.}
\label{tab:out-domain:statistics}
\end{table*}



\subsubsection*{Dataset splitting} 
Our  dataset of 5M Vietnamese translated and corrected sentence pairs is split into training, validation and test sets. We propose two splitting schemes which are \emph{in-domain} and \emph{out-of-domain}. 
For the in-domain scheme, the dataset is split based on the novel chapters, in which the first 92.5\% chapters of each novel are used for training, the next 2.5\% are for validation, and the last 5\% are for testing.  
For the out-of-domain scheme, 
we split our dataset into training, development and test sets such that no novel overlaps between them. We select novels for training, validation and test sets so that the out-of-domain data distribution is similar to the in-domain data distribution. Basic in-domain and out-of-domain data statistics are detailed in tables \ref{tab:in-domain:statistics} and  \ref{tab:out-domain:statistics}, respectively.

\section{Experimental setup}

This section presents neural MT models as well as their training details that we employ for evaluation.

\subsubsection*{Neural MT models}  
We formulate the final step of editing and polishing (i.e. post-editing) the translated sentence as a (monolingual) translation task. In particular, the translated and corrected sentences are viewed as the ones in the source and target languages, respectively. 
We employ strong neural MT models to handle the task. The first model is the well-known Transformer, in which we use its two variants of  \textbf{``Transformer-base''} and \textbf{``Transformer-large''} \cite{vaswani2017attention}. The second model is a fully convolutional model, named \textbf{``fconv''}, consisting of a convolutional encoder and a convolutional decoder \cite{gehring2017convolutional}.

\subsubsection*{Training details} For each dataset splitting scheme, we train the models on the training set using implementations from the \texttt{fairseq} library \citep{ott2019fairseq}. 
For each model, we employ the same model configuration as detailed in the corresponding paper \cite{vaswani2017attention,gehring2017convolutional}. 
We train each model with 100 epochs with the beam size of 5. We use the same shared embedding layer for both the encoder and decoder components of a neural MT model as both the translated and corrected sentences are in Vietnamese. 
We apply early stopping when no improvement is observed after 5 continuous epochs on the validation set. The model obtaining the highest BLEU score \cite{papineni2002bleu} on the validation set is then used to produce the final scores on the test set. 

We use standard MT
evaluation metrics including TER---Translation Edit Rate \cite{snover2006study}, GLEU---Google-BLEU \cite{WuSCLNMKCGMKSJL16} and BLEU, in which lower TER, higher GLEU, higher BLEU indicate better performances.

\section{Main results}
\subsubsection*{Automatic evaluation}

Table \ref{tab:results} shows in-domain and out-of-domain results for each  model as well as for the translated text. In particular, with the in-domain scheme, the neural MT models produce substantially higher GLEU and BLEU scores and a lower TER score than the translated text. This indicates that APE helps improve the quality of the translated text. Among the MT models, ``Transformer-large'' achieves the best performance with the BLEU score of 49.686 which is 1.098 and 1.753 higher than ``Transformer-base'' and ``fconv'', respectively.


\begin{table*}[!ht]
\centering
\begin{tabular}{l|l|l|l|l|l|l}
\hline
\multirow{2}{*}{Model} & \multicolumn{3}{c|}{In-domain}& \multicolumn{3}{c}{Out-of-domain} \\
\cline{2-7}
& TER$\downarrow$ & GLEU$\uparrow$ & BLEU$\uparrow$ & TER$\downarrow$ & GLEU$\uparrow$ & BLEU$\uparrow$\\
\hline
translated & 46.027 & 39.816 & 35.834 & 50.678 & 36.174 & 31.591\\ \hline
fconv & 36.539 & 49.188 & 47.933 & 43.106 & 42.654 & 40.502 \\
Transformer-base & 35.882 & 49.803 & 48.588 & 42.970 & 42.726 & 40.588 \\
Transformer-large & \textbf{35.161} & \textbf{50.763} & \textbf{49.686} & \textbf{42.892} & \textbf{42.818} & \textbf{40.704} \\ 
\hline
\end{tabular}
\caption{Experimental results on the test sets. ``translated'' denotes the result computed in using the raw translated sentence without post-editing correction. }
\label{tab:results}
\end{table*}


Regarding the out-of-domain scheme, Table \ref{tab:results} also shows a similar trend. In particular, all three neural MT models help improve the quality of the translated text with the absolute improvements of at least 7.5, 6.5, 9.0 points for TER, GLEU, BLEU, respectively. We also note that although ``Transformer-large'' consistently achieves the best TER, GLEU and BLEU scores, the out-of-domain score differences between the neural MT models are not as substantial as in the in-domain scheme. 


\subsubsection*{Human evaluation}

To better understand the performances of neural MT models, we conduct a human evaluation to manually evaluate the output quality of the three trained models. In particular, we collect a new set of 1K translated sentences which are randomly selected from 10 novels that are not in our dataset. To perform APE, we then apply each of the three  models to produce a ``corrected'' candidate output for each ``translated'' sentence, resulting in three corrected candidates.\footnote{Note that we select the 1K translated sentences to ensure that the three corrected candidates are different.}

We ask three annotators to independently vote the most suitable sentence among the translated sentence and its three corresponding  corrected candidates (here, we do not show which sentence is the translated one or corrected by which model to the annotators), thus resulting in 3,000 votes in total. The best model is ``Transformer-large'' obtaining 1,405 votes (46.8\%), compared to 815 votes (27.2\%) for ``Transformer-base'', 780 votes (26.0\%) for ``fconv'' and 0 vote for the translated sentences. We measure the inter-annotator agreements between the three annotators using Fleiss' kappa coefficient \cite{fleiss1971measuring}. The Fleiss' kappa coefficient is obtained at 0.350 which can be interpreted as \emph{fair} according to \newcite{landis1977measurement}. The results for the human evaluation are consistent with the results produced by the three models on the test sets, confirming the effectiveness of ``Transformer-large'' for APE in Vietnamese.

\section{Related work}
Our work is the first one to automatically handle the task of correcting the Vietnamese translated text of Chinese novels. However, APE is not new and has proved to be  an effective approach to handle the inaccuracies of raw MT output \cite{simard2007statistical,lagarda2009statistical,pal2016neural,nguyenetal2017sequence,correia2019simple}. 

APE approaches cover two main research directions including statistical MT-based models \cite{simard2007statistical,lagarda2009statistical} and neural MT-based models \cite{pal2016neural,correia2019simple}. In particular, \newcite{simard2007statistical} propose a statistical phrase-based MT system to post-edit the output of a rule-based MT system by handling the typical errors made by the rule-based one. Likewise, \newcite{lagarda2009statistical} utilize statistical information from a pre-trained statistical MT model to post-edit the output of another statistical MT model. \newcite{pal2016neural} propose to use Bidirectional LSTM encoder-decoder  for APE and found that it performs better than statistical phrase-based APE. \newcite{correia2019simple} present an effective APE approach where they fine-tune pre-trained BERT models \cite{devlin-etal-2019-bert} on both the BERT-based encoder and decoder.


\section{Conclusion}
We have presented the first work of APE for Vietnamese to automatically correct the Vietnamese translated text of Chinese novels. We construct the first large-scale dataset of 5M translated and corrected sentence-level pairs, extracted from 99.5K translated and corrected chapter-level pairs from 183 novels, for the Vietnamese APE task. We then compare three MT models using our dataset  under in-domain and out-of-domain data splitting schemes.  Experimental results from both the automatic and human evaluations show that the neural MT models help improve the quality of the translated text. Specifically, ``Transformer-large'' achieves the best performances w.r.t. the TER, GLEU, BLEU scores and human votes, helping to reduce the human effort in editing the translated novels, and serving as a strong model for future research and applications.
We also publicly release our dataset and model checkpoints (\textit{for research-only purpose}) at: \url{https://github.com/tienthanhdhcn/VnAPE}.

\section*{Acknowledgements}
We thank the three anonymous reviewers for their valuable comments and suggestions which help improve the quality of the paper. We would also like to thank Dat Quoc Nguyen and his team for their help and support. 

\bibliographystyle{acl_natbib}
\bibliography{REFs}

\end{document}